%% file: main.tex
\documentclass[letterpaper]{article} % DO NOT CHANGE THIS
\usepackage{aaai23}  % DO NOT CHANGE THIS
\usepackage{times}  % DO NOT CHANGE THIS
\usepackage{helvet}  % DO NOT CHANGE THIS
\usepackage{courier}  % DO NOT CHANGE THIS
\usepackage[hyphens]{url}  % DO NOT CHANGE THIS
\usepackage{graphicx} % DO NOT CHANGE THIS
\usepackage{natbib}  % DO NOT CHANGE THIS AND DO NOT ADD ANY OPTIONS TO IT
\usepackage{caption} % DO NOT CHANGE THIS AND DO NOT ADD ANY OPTIONS TO IT
\input{src/usepackage}

\usepackage{newfloat}
\usepackage{listings}
\DeclareCaptionStyle{ruled}{labelfont=normalfont,labelsep=colon,strut=off} % DO NOT CHANGE THIS
\floatstyle{ruled}
\newfloat{listing}{tb}{lst}{}
\floatname{listing}{Listing}
\title{Multi-Agent Path Finding via Tree LSTM}
\author{
    Yuhao Jiang\equalcontrib, Kunjie Zhang\equalcontrib, Qimai Li\equalcontrib, \\
    Jiaxin Chen\thanks{Corresponding author}, Xiaolong Zhu \\
}
\affiliations{
    Parametrix.AI\\
    Shenzhen, Guangdong, China\\
    \{yuhaojiang, ericzhang, qimaili, jiaxinchen, xiaolongzhu\}@chaocanshu.ai
}

\begin{document}

\maketitle

\input{src/abstract}

\input{src/sec1-introduction}

\input{src/sec2-flatland}

\input{src/sec3-approach}

\input{src/sec4-experiment}

\input{src/sec5-conclusion}

% \section{Acknowledgments}

\bibliography{aaai23}

\input{src/appendix}

\end{document}

%% file: src/usepackage.tex
\usepackage{amsmath}
\usepackage{amsfonts}

\usepackage{algorithm}
\usepackage{algorithmic}

\usepackage{booktabs}
\usepackage{multirow}

\usepackage{xcolor}
\usepackage[colorlinks = true,
            linkcolor = blue,
            urlcolor  = blue,
            citecolor = blue,
            anchorcolor = blue]{hyperref}

\usepackage{cleveref}
\crefname{figure}{Figure}{Figure}
\crefname{table}{Table}{Table}

\newcommand{\Tmax}{{T_{\mathrm{max}}}}  % max timestamp
\newcommand{\target}{{\mathrm{target}}} % target of trains, such as \target_i

\newcommand{\Xattr}{X^\mathrm{attr}}
\newcommand{\Xtree}{X^\mathrm{tree}}
\newcommand{\Hattr}{H^\mathrm{attr}}
\newcommand{\Htree}{H^\mathrm{tree}}

\newcommand{\xattr}{\mathbf{x}^\mathrm{attr}}

\newcommand{\Tree}{\mathcal{T}}
\newcommand{\Ver}{\mathcal{V}}
\newcommand{\Edge}{\mathcal{E}}
\newcommand{\ver}{\nu}
\newcommand{\x}{\boldsymbol{x}}

\newcommand{\MLP}{\operatorname{MLP}}
\newcommand{\TreeLSTM}{\operatorname{TreeLSTM}}
\newcommand{\SelfAttn}{\operatorname{Self-Attention}}

\newcommand{\Child}{\operatorname{Child}}
\newcommand{\LSTMCell}{\operatorname{LSTM-Cell}}
\newcommand{\TreeLSTMCell}{\operatorname{TreeLSTM-Cell}}

\let\oldref\ref
\renewcommand{\ref}[1]{\oldref{#1}\PackageWarning{Qimai}{\ ref is deprecated. Consider using \ cref instead.}}

\let\oldcite\cite
\renewcommand{\cite}[1]{\oldcite{#1}\PackageWarning{Qimai}{\ cite is deprecated. Consider using \ citep and \ citet instead.}}

%% file: src/abstract.tex
\begin{abstract}
In recent years, Multi-Agent Path Finding (MAPF) has attracted attention from the fields of both Operations Research (OR) and Reinforcement Learning (RL). However, in the 2021 Flatland3 Challenge, a competition on MAPF, the best RL method scored only 27.9, far less than the best OR method. This paper proposes a new RL solution to Flatland3 Challenge, which scores 125.3, several times higher than the best RL solution before. We creatively apply a novel network architecture, TreeLSTM, to MAPF in our solution. Together with several other RL techniques, including reward shaping, multiple-phase training, and centralized control, our solution is comparable to the top 2-3 OR methods. 
\end{abstract}

%% file: src/sec1-introduction.tex
\section{Introduction}\label{sec:introduction}

% MARL已经再许多地方取得巨大的成功。但是
% 介绍flatland比赛情况，参与人数、OR和RL目前的表现

%introduction大概就开头介绍mapf问题，讲讲实用价值。
%再来一段讲讲multi agent rl近年来发展飞速，取得了很多成果。
%最后一段讲讲flatland之前or和rl的performance，我们把rl的效果提升了一大截，和or comparable了

% 几个点，mapf【实用价值和challenges】
% or来做但是orblabla，rl想来做，但是目前水平 blabla
% 我们想研究这个问题，flatland作为一个instantiation来研究

% MAPF 是一个很重要的问题，有很深的现实意义，我们要研究他，但是他非常的难，总结一下challenges,目前比较好的研究方法就是or，但是or

% marl想做，在很多的东西上都取得了成功，那么自然的就希望用rl来做，但是rl目前很菜

% flatland作为一个instantiation

%我们propose了一个

Multi-agent path finding (MAPF), i.e., finding collision-free paths for multiple agents on a graph, has been a long-standing combinatorial problem. 
On undirected graphs, a feasible solution can be found in polynomial time, but finding the fastest solution is NP-hard.
And on directed graphs, even finding a feasible solution is NP-hard in general cases \citep{nebel2020computational}. Despite the great challenges of MAPF, it is of great social and economic value because many real-life scheduling and planning problems can be formulated as MAPF questions. Many operations research (OR) algorithms are proposed to efficiently find sub-optimal solutions \citep{cohen2019optimal,vsvancara2019online,ma2018multi,ma2017multi}. 
% However, despite the sub-optimal solvers, OR algorithms still suffer from low efficiency and non-real-time decisions, especially under massive-agent cases. To deal with these drawbacks, it's reminiscent of multi-agent reinforcement learning (MARL), which supports real-time decisions and efficient inference.

As an important branch in decision theory, reinforcement learning has attracted a lot of attention these years due to its super-human performance in many complex games, such as AlphaGo \citep{silver2017mastering}, AlphaStar \citep{vinyals2019grandmaster}, and OpenAI Five \citep{berner2019dota}. Inspired by the tremendous successes of RL on these complex games and decision scenarios, multi-agent reinforcement learning (MARL) is widely expected to work on MAPF problems as well. We study MARL in MAPF problems and aim to provide high-performance and scalable RL solutions. 
To develop and test our RL solution, we focus on a specific MAPF environment, Flatland.
% The existing MARL algorithms improve the inference time by a large margin compared with the OR algorithms through real-time decisions; however, they still suffer from low sample efficiency (cite), low generalization ability, and uninterpretability. 
% As advances in reinforcement learning often stems from the development of new environments that abstract problems into a form where research can be done conveniently, we focus on an instantiation of the MAPF on directed graphs, Flatland.

Flatland \citep{mohanty2020flatland,Laurent21} is a train schedule simulator developed by the Swiss Federal Railway Company (SBB). It
simulates trains and rail networks in the real world and serves as an excellent environment for testing different MAPF algorithms. Since 2019, SBB has successfully held three flatland challenges, attracting more than 200 teams worldwide, receiving thousands of submissions and over one million views. The key reasons why we focus on this environment are as follows,
\begin{itemize}
    \item \textbf{Support Massive Agents:} On the maximum size of the map, up to hundreds of trains need to be planned.
    \item \textbf{Directed Graphs and Conflicts between Agents:} Trains CANNOT move back, and all decisions are not revocable. Deadlock occurs if the trains are not well planned (\cref{fig:deadlock}), which makes this question very challenging.
    \item \textbf{Lack of High-Performance RL Solutions:} Existing RL solutions show a significant disadvantage compared with OR algorithms (27.9 vs. 141.0 scores).
\end{itemize}

To solve Flatland, we propose an RL solution by standard reinforcement learning algorithms at scale. The critical components of our RL solution are (1) the application of a TreeLSTM network architecture to process the tree-structured local observations of each agent, (2) the centralized control method to promote cooperation between agents, and (3) our optimized 20x faster feature parser. 

Our contributions can be summarized as 
(1) We propose an RL solution consisting of domain-specific feature extraction and curriculum training phases design, a TreeLSTM network to process the structured observations, and a 20x faster feature parser to improve the sample efficiency. 
(2) Our observed strategies and performance show the potential of RL algorithms in MAPF problems. We find that standard RL methods coupled with domain-specific engineering can achieve comparable performance with OR algorithm ($2^{nd}$-- $3^{rd}$ OR ). Our solution provides implementation insights to the MARL in the MAPF research community. 
(3) We open-sourced\footnotemark our solution and the optimized feature parser for further research on multi-agent reinforcement learning in MAPF problems.
\footnotetext{https://github.com/liqimai/flatland-marl}
% Multi-agent path finding (MAPF) is to find collision-free paths for multiple agents on a graph. It has been a long-standing combinatorial problem. Finding the fastest solution for MAPF is still an open problem because it is NP-hard. On undirected graphs, a feasible solution can be found in polynomial time, but on directed graphs, even finding a feasible solution is NP-hard in general case \citet{nebel2020computational}.

% Flatland environment \citep{mohanty2020flatland,Laurent21} ...

%% file: src/sec2-flatland.tex
\section{Flatland3 Environment}\label{sec:environment}

Flatland is a simplified world of rail networks in which stations are connected by rails. Players control trains to run from one station to another. Its newest version, Flatland3, consists of the following rules.

\begin{itemize}
    \item The time is discretized into timestamps from 0 to $\Tmax$. 
    
    \item There are $N$ trains and several cities. Trains are numbered from $1$ to $N$.
    
    \item Trains' action space consists of five actions, \{do\_nothing, forward, stop, left, right\}. Trains are not allowed to move backward and must go along the rails. 
    
    \item Trains have different speeds. Each train $i$ has its own speed $s_i$, and can move one step every ${1}/{s_i}$ turns. The time cost of one step ${1}/{s_i}$ is guaranteed to be an integer, and the largest possible speed is 1, i.e., one step a turn.
    
    \item For each train $i$, it has an earliest departure time $A_i$ and a latest arrival time $B_i$. Each train can depart from its initial station only after its earliest departure time $A_i$ and should try its best to arrive at its target station before the latest arrival time $B_i$.
    
    \item Trains randomly break down (malfunction) while running or waiting for departure. After the breakdown, the train must stay still for a period of time before moving again.
\end{itemize}

\begin{figure}[tb]
    \centering
    \includegraphics[width=0.45\textwidth]{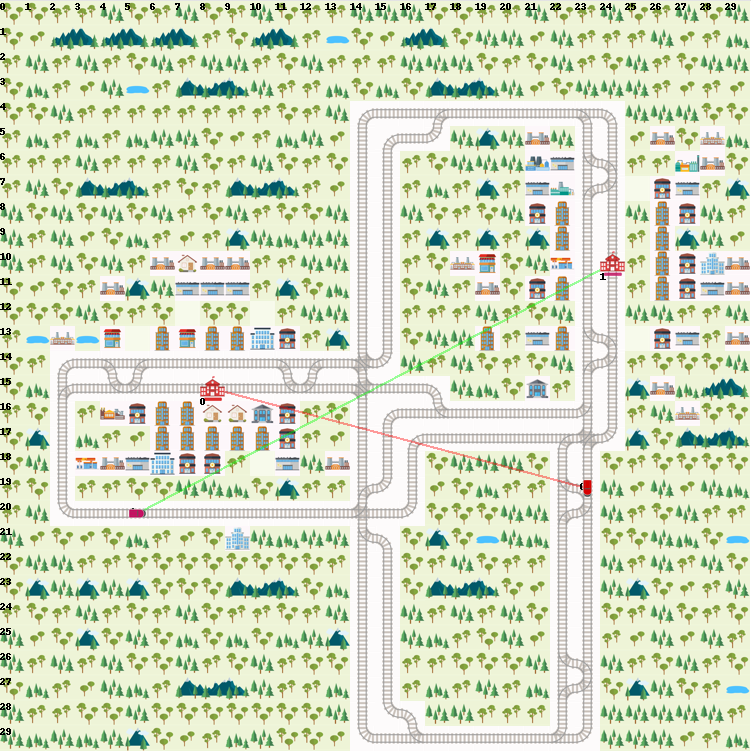}
    \caption{A $30\times30$ flatland map. There are two stations and two trains. The lines connecting trains and stations indicate trains' targets.}
    \label{fig:flatland}
\end{figure}

\input{image/rail/rail}

\paragraph{Reward} 
The goal is to control all trains to reach target stations before their latest arrival time $B_i$. Every train will get a reward $R_i$ in the end. $T_i$ denotes the arrival time of each train. 
\begin{itemize}
    \item If a train arrives on time, then it scores 0. 
    \item If it arrives late, it gets a negative reward $B_i - T_i$ as a penalty, according to how late it is. 
    \item If a train does not manage to arrive at its target before the end time $\Tmax$, then the penalty consists of two parts, a temporal penalty and a spatial penalty. The temporal penalty is $B_i-\Tmax$, reflecting how late it is. The spatial penalty is decided by the shortest path distance $d_i$ between its final location at time $\Tmax$ and its target. 
\end{itemize}
Formally, $R_i$ is defined as
\begin{equation}
    R_i = \left\{
    \begin{aligned}
        & 0,          && \text{if}~ T_i \le B_i; \\
        & B_i - T_i,  && \text{if}~ B_i < T_i \le \Tmax; \\
        & B_i - \Tmax - d_i^{(\Tmax)}, && \text{if}~ T_i > \Tmax;
    \end{aligned}
    \right.
\end{equation}
where $d_i^{(\Tmax)}$ is the distance between train $i$ and its target at time $\Tmax$,
\begin{equation}
    d_i^{(t)} = d\left((x_i^{(t)}, y_i^{(t)}), \;\target_i\right).
\end{equation}
Our goal is to maximize the sum of individual rewards
\begin{equation}
    R = \sum_{i=0}^{N} R_i.
\end{equation}
Apparently, $R$ is always non-positive, and $R=0$ if and only if all trains reach targets on time. $\lvert R\lvert$ can be arbitrarily large, as long as the map size is sufficiently large and the algorithm performance is sufficiently bad.

\paragraph{Normalized Reward}
The magnitude of total reward $R$ greatly relies on the problem scale, such as the number of trains, the number of cities, the speeds of trains, and map size. To make rewards of different problem scales comparable, they are normalized as follows:
\begin{equation}
    \bar{R} = 1 + \frac{R}{N\Tmax},
\end{equation}
where $N$ is the number of trains. The environment generating procedure guarantees $\bar{R} \in [0, 1]$ by adjusting $\Tmax$. Normalized reward serves as the standard criterion for testing algorithms.

%% file: image/rail/rail.tex
\begin{figure}[tb]
    \centering
    \def\arraystretch{1.0}
    \setlength{\tabcolsep}{0.2em}
    \begin{tabular}{|l|l|l|l|l|}
    \hline
        \includegraphics[width=0.08\textwidth]{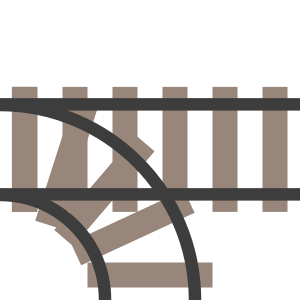} &
        \includegraphics[width=0.08\textwidth]{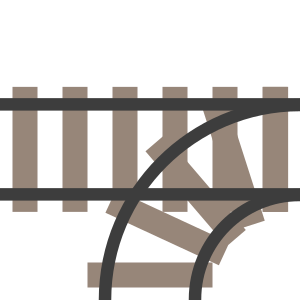} &
        \includegraphics[width=0.08\textwidth]{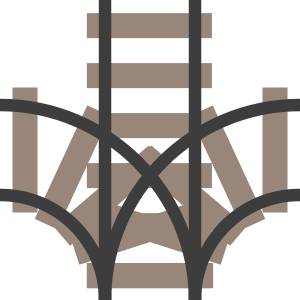} & 
        \includegraphics[width=0.08\textwidth]{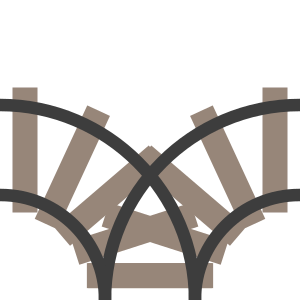} &
        \includegraphics[width=0.08\textwidth]{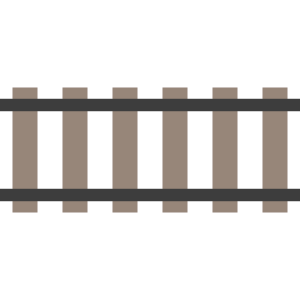}
        \\
    \hline
        \includegraphics[width=0.08\textwidth]{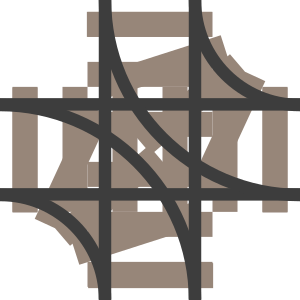} &
        \includegraphics[width=0.08\textwidth]{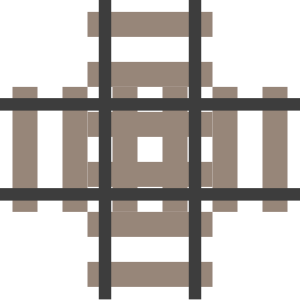} &
        \includegraphics[width=0.08\textwidth]{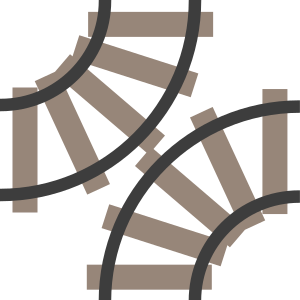} &
        \includegraphics[width=0.08\textwidth]{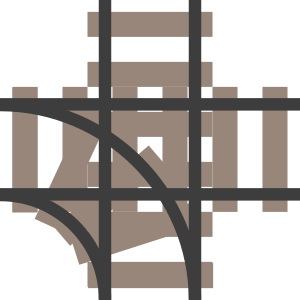} &
        \includegraphics[width=0.08\textwidth]{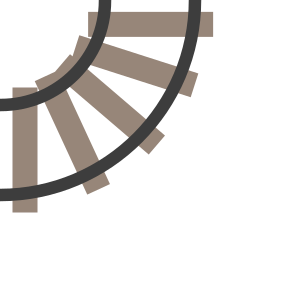}
        \\
    \hline
    \end{tabular}
    \caption{Different types of rail cells.}
    \label{fig:rails}
\end{figure}

%% file: src/sec3-approach.tex
\section{Our Approach}\label{sec:approach}
The RL solution we provide is cooperative multi-agent reinforcement learning. Each agent independently observes a part of the map localized around itself and encodes the part of the map topology into tree-structured features. 
Neural Networks independently process each agent's observation in the first several layers by TreeLSTM, then followed by several self-attention blocks to encourage communications between agents so that a train is able to be aware of others' local observations and forms its own knowledge of the global map. Rewards are shared by all agents to promote cooperation between them. Our network is trained by Proximal Policy Optimization (PPO) \citep{Schulman2017ProximalPO} algorithm.

\subsection{Feature Extraction}
The extracted features consist of two parts, $\Xattr$ and $\Xtree$. 

\paragraph{Agent Attributes} The first part, $\Xattr=\{\xattr_i\}_{i=1}^N$, are attributes of each agent, such as ID, earliest departure time, latest arrival time, their current state, direction, and the time left, etc. See \cref{tab:agent attribute} for detailed contents of $\Xattr$.

\input{table/feature_table}

\paragraph{Tree Representation of Possible Future Paths} The second part $\Xtree$ is the main part of the observations. It encodes possible future paths of each agent as well as useful information about these paths into a tree-like structure. We take the rail network as a directed graph and construct a spanning tree for each agent by a depth-limited BFS (breadth-first-search) starting from its current location. Each node in the tree represents a branch the agent may choose, see \cref{fig:tree}. Formally, the spanning tree we construct for each train $i$ is $\Tree_i= (\Ver_i, \Edge_i)$ with node set $\Ver_i$ and edge set $\Edge_i$. Each node $\ver\in\Ver_i$ is associated with a vector $\x_\ver$, containing useful information about this branch. See \cref{tab:node feature} for detailed contents of $\x_\ver$. So, $\Xtree$ contains both the tree structure and these associated node features:
\begin{equation}
    \Xtree = \{(\Tree_i, \Xtree_i)\}_{i=1}^N,
\end{equation}
where $\Xtree_i = \{\x_\ver| \ver\in\Ver_i\}$.

Such tree representation is provided by Flatland3 environment and has also been explored by other RL methods \citep{mohanty2020flatland, Laurent21}. However, no RL method before has achieved comparable performance as ours because we made the following improvements.
\begin{itemize}
    \item First, all methods before concatenate node features together into a long vector so that it can be fed into MLPs. Normal networks can only process vectors, not tree-like input. After concatenation, the underlying structures of trees are lost. In contrast, we think the tree structures are super important for decision-making and must be preserved, as they encode map topology. In \cref{sec:netwrok}, we processing such tree-structured data by a special neural network structure, TreeLSTM \cite{tai2015improved}.
    \item Second, our trees are much deeper than others. Tree depth decides the range of agents' field of view and thus significantly affects the performance. Extracting tree representation is a computationally intensive task. The flatland3 built-in implementation of tree representation is super slow because of the poor efficiency of Python language and the unnecessary complete ternary tree it uses, so the RL methods before have a very limited tree depth, typically 3. We re-implement tree construction by C++ and prune complete ternary trees into normal trees. Our implementation is 20x faster than the built-in one and enables us to build trees with depths of more than 10.
    \item Third, we build trees in the BFS manner, while the built-in implementation is in the DFS manner. Constructing a spanning tree in a DFS manner makes some nodes near the root on the graph become far from the root, which is a disadvantage.
\end{itemize}

\subsection{Neural Network Architecture} \label{sec:netwrok} % 或者给模型起个名字? 

\begin{figure}[t]
    \centering
    \includegraphics[width=0.4\textwidth]{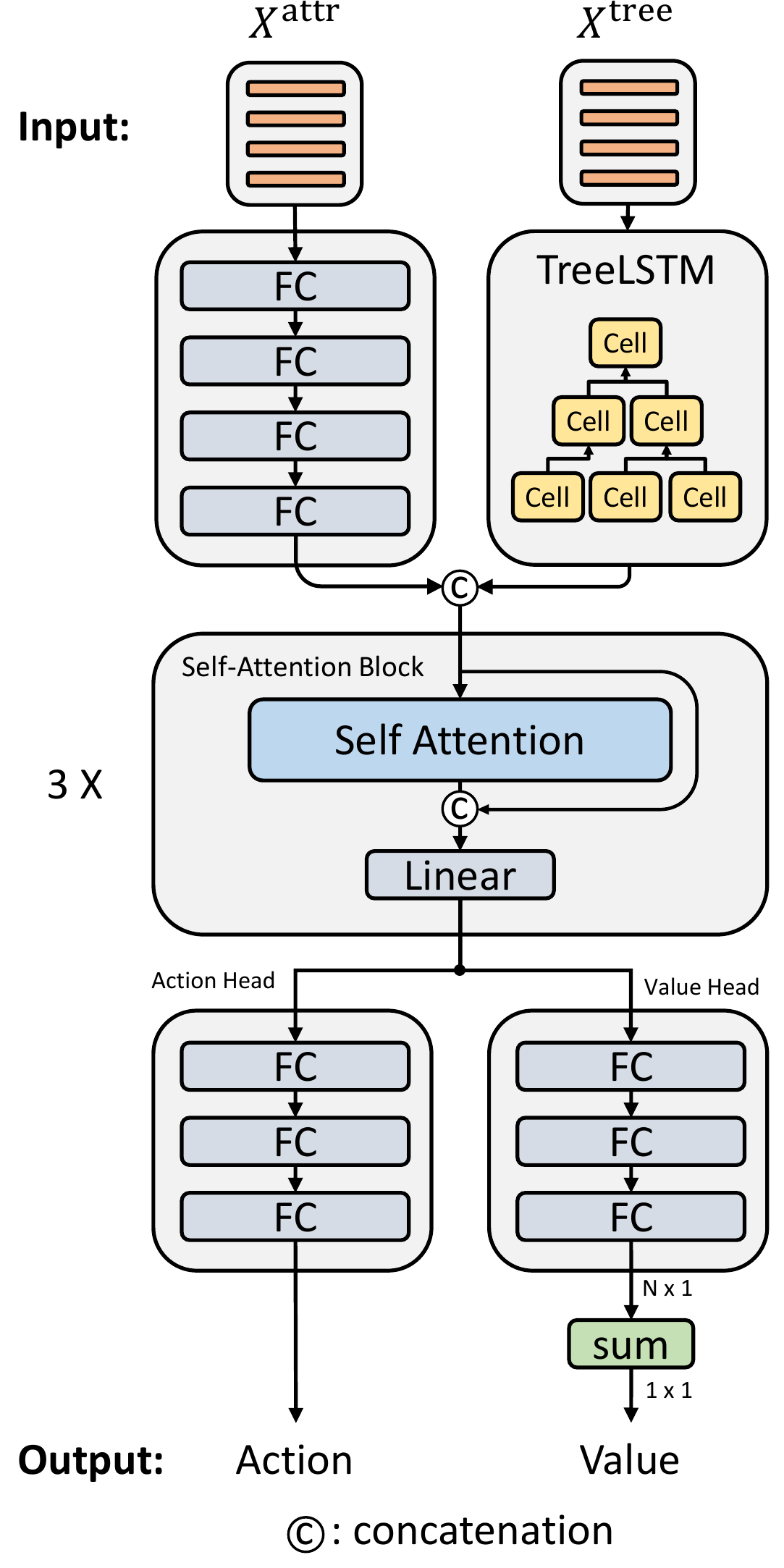}
    \caption{Overview of our network architecture. FC stands for fully connected layer.}
    \label{fig:network}
\end{figure}

As shown in \cref{fig:network}, our neural networks first process $\Xattr$ by a 4-layer MLP and process $\Xtree$ by TreeLSTM \citep{tai2015improved}. TreeLSTM is a variant of LSTM designed for tree-structured data, whose details will be elaborated later.
\begin{align}
    \Hattr &= \MLP(\Xattr) \\
    \Htree &= \TreeLSTM(\Xtree)
\end{align}

Then, we concatenate $\Hattr$ and $\Htree$ together and feed them into three consecutive self-attention blocks to encourage communications between agents. With the self-attention mechanism \citep{vaswani2017attention}, a train is able to be aware of other trains' observations and forms its own knowledge of the global map. 
\begin{align}
    H^{(0)} & = [\Hattr, \Htree], \\
    H^{(l)} & = \SelfAttn\left(H^{(l-1)}\right), \quad l=1,2,3.
\end{align}

Finally, $H^{(3)}$ is fed into two different heads to obtain final actions logits $A\in\mathbb{R}^{N\times 5}$ and estimated state-value $v\in\mathbb{R}$.
\begin{align}
    A &= \MLP(H^{(3)}) \\
    V &= \MLP(H^{(3)}) \\
    v &= \textstyle \sum_{i=0}^N V_i
\end{align}

\paragraph{TreeLSTM} LSTM \citep{hochreiter1997long}, as a kind of RNN, was designed to deal with sequential data. Each LSTM cell takes state $(c_{t-1}, h_{t-1})$ of last cell and a new $x_t$ as input, and output new cell state $c$ and $h$ to next cell.
\begin{equation}
    (c_t, h_t) = \LSTMCell\left(x_t, (c_{t-1}, h_{t-1})\right) \\
\end{equation}

Sequential data is a special case of trees, where each node has a unique child --- its successor. \citeauthor{tai2015improved} modified its structure to deal with general trees in \citeyear{tai2015improved}. Tree differs from sequential data in the allowed number of children. Unlike sequential data, nodes in a tree are allowed to have multiple children. As a result, TreeLSTM receives a set of children's output as input instead:
\begin{align}
    & (c_t, h_t) = \TreeLSTMCell(x_t,  S_t), 
\end{align}
where $S_t = \{(h_k, c_k)\;|\;k\in\Child(t)\}$. Within TreeLSTM cells, there are several ways to aggregate children's states, leading to different variants of TreeLSTM. In our network, we adopt Child-sum TreeLSTM. See \citet{tai2015improved} for detailed structures of TreeLSTM cells.

% \begin{align}
%     i_t &= \sigma(W^{(i)} x_t  + U^{(i)} h_{t-1} + b^{(i)}), \\
%     f_t &= \sigma(W^{(f)} x_t  + U^{(f)} h_{t-1} + b^{(f)}), \\
%     o_t &= \sigma(W^{(o)} x_t  + U^{(o)} h_{t-1} + b^{(o)}), \\
%     u_t &= \tanh(W^{(u)} x_t  + U^{(u)} h_{t-1} + b^{(u)}), \\
%     c_t &= i_t \odot u_t + f_t \odot c_{t-1}, \\
%     h_t &= o_t \odot \tanh(c_t),
% \end{align}

% \begin{align}
%     \coloremph{\tilde{h}_t} & \coloremph{= \sum_{k\in\Child(t)} h_k}\\
%     i_t &= \sigma(W^{(i)} x_t  + U^{(i)} \coloremph{\tilde h_t} + b^{(i)}), \\
%     \coloremph{f_{tk}} &= \sigma(W^{(f)} x_t  + U^{(f)} \coloremph{h_k} + b^{(f)}), \quad \coloremph{k\in\Child(t)}, \\
%     o_t &= \sigma(W^{(o)} x_t  + U^{(o)} \coloremph{\tilde h_t} + b^{(o)}), \\
%     u_t &= \tanh(W^{(u)} x_t  + U^{(u)} \coloremph{\tilde h_t} + b^{(u)}), \\
%     c_t &= i_t \odot u_t + \coloremph{\sum_{k\in\Child(t)} f_{tk} \odot c_k}, \\
%     h_t &= o_t \odot \tanh(c_t),
% \end{align}

\subsection{Reward Design}
Agents are given rewards at every time step, according to their performance within the moment. Besides the normalized reward generated by the environment, agents are also rewarded when they depart from stations, arrive at targets, and get penalized when deadlocks happen. To promote cooperation between them, these rewards are shared by all agents, and no credit assignment is performed. As a result, a single agent is encouraged to wait for others if the waiting can lead to global efficiency improvement.

% The reward of the model is the sum of normalized environmental reward, arrival reward, departure reward, and deadlock penalty, which are all differences between time steps and illustrated below.

\paragraph{Environmental Reward}
Agents are rewarded environmental reward $r^{(e)}_t$ at time step $t$:
\begin{equation}
    r^{(e)}_t = \bar{R}_t,
\end{equation}
where $\bar{R}_t$ is the normalized environmental reward agents get in time step $t$.

\paragraph{Departure Reward}
We reward agents when there are new agents departing:
\begin{equation}
    r^{(d)}_t = \frac {n^{(d)}_t - n^{(d)}_{t - 1}} {N},
\end{equation}
where $n^{(d)}_t$ is the number of agents departing at or before time step $t$.

\paragraph{Arrival Reward}
We reward agents when there are new arrivals:
\begin{equation}
    r^{(a)}_t = \frac {n^{(a)}_t - n^{(a)}_{t - 1}} {N},
\end{equation}
where $n^{(a)}_t$ is the number of arrival so far at time step $t$.

\begin{figure}[t]
    \centering
    \includegraphics[width=0.35\textwidth]{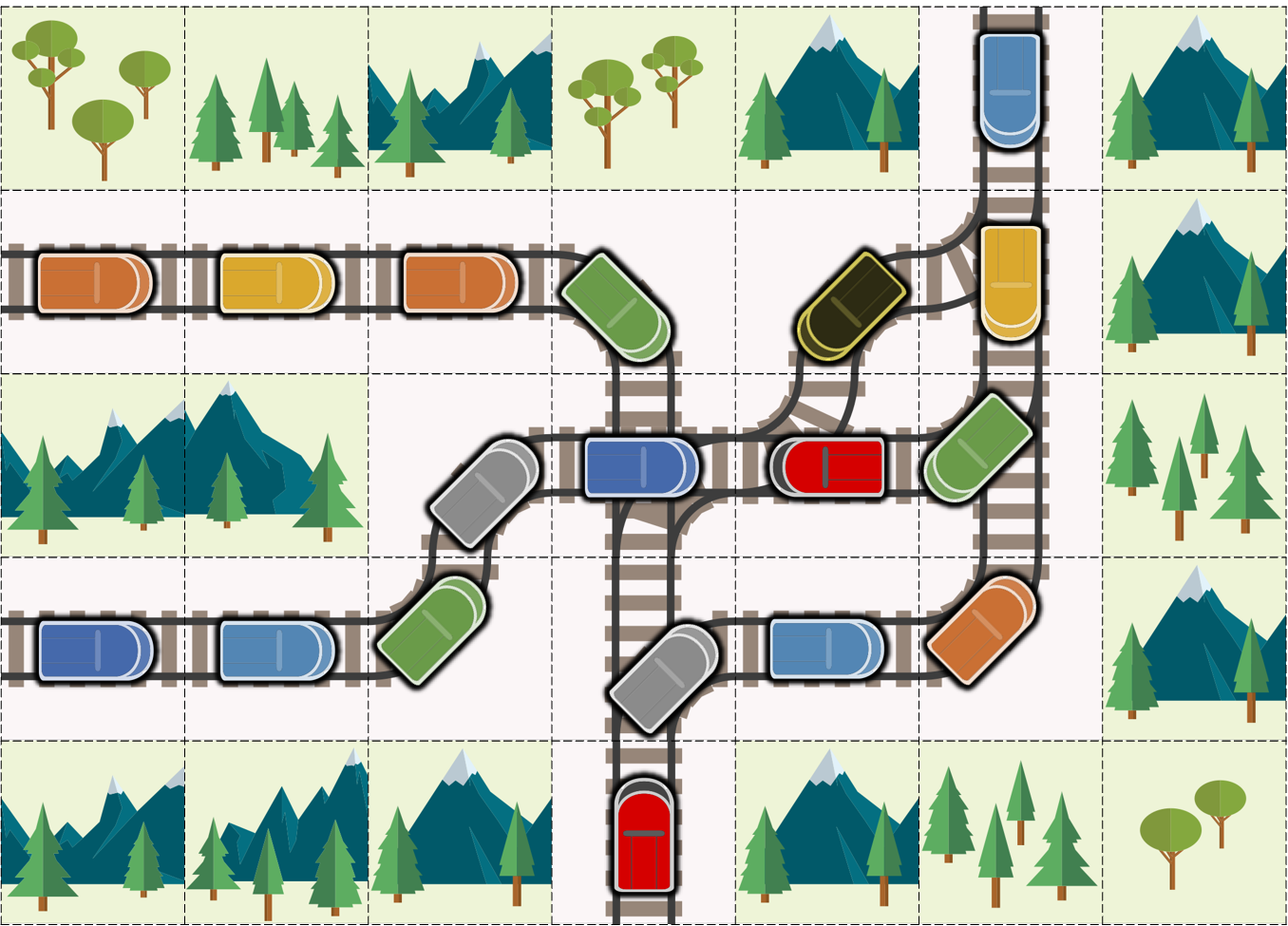}
    \caption{Because trains are not allowed to go backward, if two trains go into a single rail in opposite directions, a deadlock happens.}
    \label{fig:deadlock}
\end{figure}

\paragraph{Deadlock Penalty}
Because trains are not allowed to go backward, if two trains go into a single rail in opposite directions, a deadlock happens and no train can pass this rail again (see \cref{fig:deadlock}). So, we give a penalty when new deadlocks happen:
\begin{equation}
    r^{(l)}_t = \frac {n^{(l)}_t - n^{(l)}_{t - 1}} {N},
\end{equation}
where $n^{(l)}_t$ is the number of deadlocks on the map at time step $t$.

\paragraph{Total Reward}
The final reward we give to agents at time $t$ is a weighted sum of all terms above:
\begin{equation}
    r_t = c_e r_t^{(e)} + c_a r_t^{(a)} + c_d r_t^{(d)} - c_l r_t^{(l)},
\end{equation}
where $c_e,c_a,c_d,c_l$ are weight parameters.

%% file: table/feature_table.tex
\begin{table}[t]
    \centering
    \begin{tabular}{llll}
    \toprule
    & Name & Dim. & Type\\
    \midrule
    \multirow{5}{*}{Timetable}
    & Train ID & 1 & int \\
    & Earliest departure time & 1 & int \\
    & Latest arrival time & 1 & int \\
    & Initial direction &  4 & one-hot \\
    & Initial distance to target & 1 & int\\
    
    \midrule
    \multirow{7}{*}{Spatial}
    & Road type & 11 & one-hot \\
    & Possible transitions & 16 & binary \\
    & Current direction & 4 & one-hot \\
    & Last-turn direction & 4 & one-hot\\
    & Deadlocked or not & 1 & binary\\
    & Valid actions & 5 & binary \\
    & Distance to target & 1 & int \\
    
    \midrule
    \multirow{3}{*}{Temporal}
    & Current time & 1 & int \\
    & \#turns before late & 1 & int \\
    & Arrival time & 1 & int \\
    
    \midrule
    \multirow{8}{*}{State}
    & State &  7 & one-hot \\
    & Is off-map state & 1 & binary\\
    & Is on-map state & 1 & binary\\
    & Is malfunction state & 1 & binary \\
    & Is moving or not & 1 & binary\\
    & Malfunction ends & 1 & binary \\
    & Left malfunctional turns & 1 & int\\
    & Speed state-machine & 5 & int\\
    \bottomrule
    \end{tabular}
    \caption{Extracted agent attributes in $\Xattr$.}
    \label{tab:agent attribute}
\end{table}

\begin{figure}[t]
    \centering
    \includegraphics[width=0.47\textwidth]{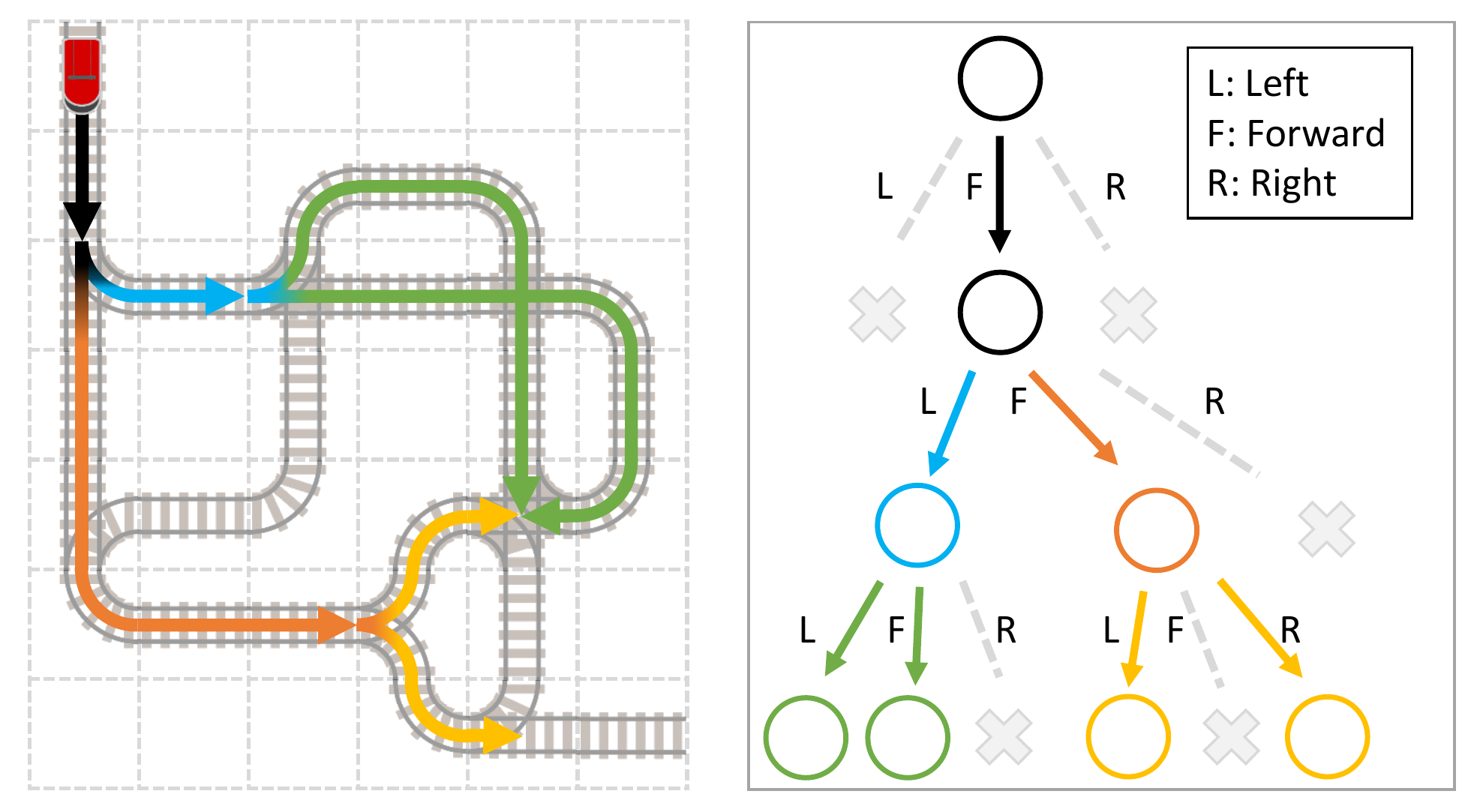}
    \caption{Construct a spanning tree for each agent to encode its possible future paths. Each node represents a branch the agent may choose.}
    \label{fig:tree}
\end{figure}

\begin{table}[t]
    \centering
    \begin{tabular}{lll}
    \toprule
    Description & Dim. & Type\\
    \midrule
    \#agents in same direction & 1 & int \\
    \#agents in opposite direction & 1 & int \\
    \#agents ready to depart & 1 & int \\
    \#agents in malfunction & 1 & int \\
    Distance to agent's own target & 1 & int \\
    Distance to other agents' targets, if any & 1 & int \\
    Distance to other agents, if any & 1 & int \\
    Distance to potential conflict, if any & 1 & int \\
    Distance to unusable switch, if any & 1 & int \\
    Branch length & 1 & int \\
    % Distance to target \\
    Slowest speed of observed agents & 1 & float\\
    \bottomrule
    \end{tabular}
    \caption{Useful information extracted for each node in spanning tree.}
    \label{tab:node feature}
\end{table}

%% file: src/sec4-experiment.tex
\section{Experiments}\label{sec:experiment}

\input{table/experiments_info}
\input{table/compare_phase_1_2}
\input{table/generalization_up}

\subsection{Experiment Settings}
We largely followed the final round (round 2) configurations of the Flatland3 challenge to conduct experiments so that our results are comparable with the ones on the challenge leaderboard \footnotemark. There are 15 test stages in the final round, and each stage contains 10 test cases. Problem scales (\cref{tab:main-result}) and difficulty gradually increase from initial stages to advanced stages. The first stage is the smallest one with 7 agents on a $30\times30$ map, while the last stage contains 425 agents on a $158\times158$ map. Teams' submissions are tested stage by stage. A team can proceed to the next stage only if they pass the last stage (arrival ratio reaches 25\%).

\footnotetext{\url{https://www.aicrowd.com/challenges/flatland-3/leaderboards}}

\input{table/main-result}
\input{table/leaderboard}
\input{table/generalization_down}

\subsection{Multiple Phase Training}
To reduce training difficulty, we train our models in curriculum learning style, and the whole training process can be roughly divided into three phases. In phases I and II, we train a model in 50-agent environments. We found that the learned model generalizes well to smaller environments but not larger ones. In phase III, models are initialized by the one learned in phase II and fine-tuned in settings with more agents.

\paragraph{Phase I}
Initially, we only use the environmental reward, arrival reward, and deadlock penalty to encourage the trains to march on their targets and avoid deadlocks (see Phase-I in \cref{tab:experiments}). 
After training, 70\% agents in 50-agent environments can reach their targets, and a normalized reward of 0.859 is achieved. However, 21\% agents have never departed because of the deadlock penalty. Agents choose not to depart and behave conservatively to avoid being penalized by deadlocks.

\paragraph{Phase II}
In phase I, many trains do not depart, so we add a departure reward to encourage more departures (see Phase-II in \cref{tab:experiments}). 
The experiment is initialized by the phase-I model, and after 5 days of training, its arrival ratio increases to 86.2\%, and almost all the trains have departed (\cref{tab:phase_1_and_2}).

\paragraph{Phase III} 
Finally, we deal with environments with more than 80 agents. Training models in so large environments from scratch is very difficult, so we adopt curriculum learning. Models for large environments are initialized by parameters learned in small environments. Although models learned in small environments are able to directly generalize to large environments (\cref{tab:generaliztion_up}), fine-tuning in large environments increases performances significantly.

\subsection{Results and Analysis}
Our stage-specific results are reported in \cref{tab:main-result}; Final scores as well as top 10 teams' scores in Flatland3 challenge are listed in \cref{tab:leaderboard}. In summary, we scored 125.3, ranking top 2--3 on the leaderboard, while the best RL method before scored only 27.9. More specifically, we observe the following phenomena:
\begin{itemize}
    \item No RL method before us managed to pass Test\_03 stage, while our method passed all 15 stages.
    \item While the number of agents increases, model performance decreases, which suggests large-scale problems are more difficult than we expected. 
    \item When the numbers of agents are equal (Test\_04 to Test\_08), model performance increases with a larger map and more cities because it leads to lower agent density and less traffic congestion.
    \item Compared to the third-best team, we achieved a higher environmental reward but a lower arrival ratio. This indicates that environmental rewards are not always consistent with arrival ratios because arrival ratios only care agents arrive or not while environmental rewards also care about how fast agents arrive. They are two highly related but different objectives. Similar phenomena can be observed in team  An\_Old\_Driver and team Zain. 
\end{itemize}

\paragraph{Generalization across environment scales} 
We are also interested in the generalization ability of our models and particularly interested in the generalization across environment scales. As \cref{tab:generaliztion_up} shows, models learned in small environments are able to generalize to large environments but perform worse than the one fine-tuned in large environments. \cref{tab:generalization_down} shows that models specialized in large environments are also able to generalize to small environments but perform worse than the ones learned in small environments. To achieve optimal performance, we need to train multiple scale-specific models. 

\paragraph{Agent Cooperation}
We observed many self-organized cooperative patterns in agents' behaviors. They learn to line up to march in a compact manner (\cref{fig:lineup}). Fast ones learned to overtake the slow ones (\cref{fig:overtake}). Slow trains make way for fast ones (\cref{fig:makeway}). When there are two parallel rail lanes, trains spontaneously line up as if they are in a two-way street (\cref{fig:parallel}).

%% file: table/experiments_info.tex
\begin{table}[t]
    \centering
    \small
    \setlength{\tabcolsep}{3pt}
    \begin{tabular}{lrrrlrl}
        \toprule
        % Experiment
        & \multicolumn{1}{c}{\#agents} 
        % & \multicolumn{1}{c}{Tree Depth} 
        & \multicolumn{4}{c}{Reward Weights} 
        & \multicolumn{1}{c}{Initialized by} \\
        \cmidrule(lr){3-6}
         & & $c_e$ & $c_a$ & $c_d$ & $c_l$& \\
        \midrule
        Phase-I & 50 & 1 & 5 & 0 & 2.5 & N/A \\
        Phase-II & 50 & 0 & 5 & 1 & 2.5 & Phase-I \\
        Phase-III-50 & 50 & 0 & 5 & 1 & 2.5 & Phase-II \\
        Phase-III-80 & 80 & 1 & 5 & 0.1 & 2.5 & Phase-III-50 \\
        Phase-III-100 & 100 & 1 & 5 & 0.1 & 2.5 & Phase-III-50 \\
        Phase-III-200 & 200 & 1 & 5 & 0.1 & 2.5 & Phase-III-100 \\
        \bottomrule
    \end{tabular}
    
    \caption{Settings of different phases.}
    \label{tab:experiments}
\end{table}

%% file: table/compare_phase_1_2.tex
\begin{table}[t]
    \centering
    \begin{tabular}{lrrr}
        \toprule
                & Arrival\% & Env. Reward & Depart\% \\
        \midrule
        Phase-I & 70.0 & 0.859 & 79.0 \\
        Phase-II & 86.2 & 0.920 & 99.2 \\
        \bottomrule
    \end{tabular}
    
    \caption{Compare Phase-I and Phase-II.}
    \label{tab:phase_1_and_2}
\end{table}

%% file: table/generalization_up.tex
\begin{table}[t!]
    \centering
    \newcommand{\stdsize}{\small}
    \small
    \setlength{\tabcolsep}{3pt}
    \begin{tabular}{lllll}
        \toprule
        Model & \multicolumn{2}{c}{Phase-III-50} & \multicolumn{2}{c}{Phase-III-80} \\
        \cmidrule(lr){2-3} \cmidrule(lr){4-5}
        Test Stage & 
        \multicolumn{1}{c}{Arrival\%} & 
        Env. Reward & 
        \multicolumn{1}{c}{Arrival\%} & 
        Env. Reward \\
        \midrule
        \tt{Test\_04} & 50.5 \stdsize $\pm$19.7 & .781 \stdsize $\pm$.079 & 62.6 \stdsize $\pm$11.9 & .812 \stdsize $\pm$ .051 \\
        \tt{Test\_05} & 49.4 \stdsize $\pm$21.0 & .779 \stdsize $\pm$.073 & 62.9 \stdsize $\pm$12.8 & .824 \stdsize $\pm$ .049 \\
        \tt{Test\_06} & 51.6 \stdsize $\pm$20.4 & .788 \stdsize $\pm$.083 & 70.6 \stdsize $\pm$ 6.2 & .859 \stdsize $\pm$ .028 \\
        \tt{Test\_07} & 52.2 \stdsize $\pm$20.2 & .803 \stdsize $\pm$.086 & 65.4 \stdsize $\pm$12.6 & .833 \stdsize $\pm$ .051 \\
        \tt{Test\_08} & 52.9 \stdsize $\pm$17.9 & .789 \stdsize $\pm$.083 & 74.3 \stdsize $\pm$ 9.6 & .877 \stdsize $\pm$ .029 \\
        \bottomrule
    \end{tabular}
    \caption{Phase-III-50 is trained in 50-agent environments, but Test\_04 to Test\_08 are 80-agent environments. Model Phase-III-80 is initialized by Phase-III-50, and fine-tuned in 80-agent environments. After fine-tuning, both arrival ratios and environmental rewards increase significantly.}
    \label{tab:generaliztion_up}
\end{table}

%% file: table/main-result.tex
\begin{table*}[t]
    \centering
    \begin{tabular}{llrcrlrr}
        \toprule
        Test Stage & 
        \multicolumn{1}{c}{Model} & 
        \#agents & 
        Map Size & 
        \#cities & 
        \multicolumn{1}{c}{Arrival\%} & 
        Env. Reward & 
        Avg. Time/s \\
        \midrule
        \tt{Test\_00} & Phase-III-50 & 7 & $30 \times 30$ & 2 & $94.3 \pm 10.0$ & $.957 \pm .030$ & 7.022 \\
        \tt{Test\_01} & Phase-III-50 & 10 & $30 \times 30$ & 2 & $92.0 \pm 9.2$ & $.947 \pm .047$ & 8.430 \\
        \tt{Test\_02} & Phase-III-50 & 20 & $30 \times 30$ & 3 & $87.0 \pm 13.6$ & $.934 \pm .063$ & 16.486 \\
        \tt{Test\_03} & Phase-III-50 & 50 & $30 \times 35$ & 3 & $86.2 \pm 10.2$ & $.922 \pm .047$ & 32.292 \\
        \tt{Test\_04} & Phase-III-80 & 80 & $35 \times 30$ & 5 & $62.6 \pm 11.9$ & $.812 \pm .051$ & 40.580 \\
        \tt{Test\_05} & Phase-III-80 & 80 & $45 \times 35$ & 7 & $62.9 \pm 12.8$ & $.824 \pm .049$ & 60.009 \\
        \tt{Test\_06} & Phase-III-80 & 80 & $40 \times 60$ & 9 & $70.6 \pm 6.2$ & $.859 \pm .028$ & 99.566 \\
        \tt{Test\_07} & Phase-III-80 & 80 & $60 \times 40$ & 13 & $65.4 \pm 12.6$ & $.833 \pm .051$ & 109.386 \\
        \tt{Test\_08} & Phase-III-80 & 80 & $60 \times 60$ & 17 & $74.3 \pm 9.6$ & $.877 \pm .029$ & 160.928 \\
        \tt{Test\_09} & Phase-III-100 & 100 & $80 \times 120$ & 21 & $59.7 \pm 15.7$ & $.795 \pm .067$ & 480.971 \\
        \tt{Test\_10} & Phase-III-100 & 100 & $100 \times 80$ & 25 & $57.6 \pm 16.7$ & $.779 \pm .067$ & 346.861 \\
        \tt{Test\_11} & Phase-III-200 & 200 & $100 \times 100$ & 29 & $52.8 \pm 5.8$ & $.790 \pm .033$ & 488.549 \\
        \tt{Test\_12} & Phase-III-200 & 200 & $150 \times 150$ & 33 & $57.3 \pm 5.0$ & $.777 \pm .037$ & 1314.509 \\
        \tt{Test\_13} & Phase-III-200 & 400 & $150 \times 150$ & 37 & $34.9 \pm 7.2$ & $.704 \pm .031$ & 2029.058 \\
        \tt{Test\_14} & Phase-III-200 & 425 & $158 \times 158$ & 41 & $39.3 \pm 9.7$ & $.721 \pm .038$ & 2329.925 \\
        \bottomrule
    \end{tabular}
    \caption{Performance of our models in 15 test stages.}
    \label{tab:main-result}
\end{table*}

%% file: table/leaderboard.tex
\begin{table}[t]
    \centering
    \begin{tabular}{rllrr}
        \toprule
        Rank & 
        \multicolumn{1}{c}{Team} & 
        \multicolumn{1}{c}{Tag} & 
        \multicolumn{1}{c}{Score} & 
        \multicolumn{1}{c}{Arv.\%}  \\
        \midrule
        1 & \small \tt{An\_Old\_Driver}    & OR    & 141.0 & 88.0 \\
        2 & \small \tt{Zain}               & OR    & 132.5 & 88.8 \\
        - & Ours                           & RL    & 125.3 & 66.4      \\
        3 & \small \tt{SmartTrains}        & OR    & 118.0 & 76.9 \\
        4 & \small \tt{dsa}                & OR    & 107.5 & 44.2 \\
        5 & \small \scriptsize \tt{stavros\_kakoulidis}& other & 40.5 & 55.6 \\
        6 & \small \tt{UniTeam}            & other & 29.9 & 39.1 \\
        7 & \small \tt{WaveTeam}           & RL    & 27.9 & 38.6 \\
        8 & \small \tt{SOBA}               & RL    & 27.8 & 32.6 \\
        9 & \small \tt{ChewChewChew}       & RL    & 20.0 & 30.6 \\
        10& \small \tt{fridayPhenom}       & OR    &  6.9 & 18.6 \\
        \bottomrule
    \end{tabular}
    \caption{Compare our results with the top 10 in Flatland3 challenge leaderboard. The scores are collected by accumulating environmental rewards from all 15 test stages.}
    \label{tab:leaderboard}
\end{table}

%% file: table/generalization_down.tex
\begin{table}[t]
    \centering
    \begin{tabular}{crll}
        \toprule
        Test Stage & 
        \#agents &
        \multicolumn{1}{c}{Arrival\%} & 
        Env. Reward \\
        \midrule
        \tt{Test\_03} & 50 & 56.4 $\pm$ 10.8 & .815 $\pm$ .038  \\
        \tt{Test\_04} & 80 & 50.7 $\pm$ 9.6 & .810 $\pm$ .023 \\
        \tt{Test\_05} & 80 & 47.4 $\pm$ 9.0 & .794 $\pm$ .021 \\
        \tt{Test\_06} & 80 & 53.0 $\pm$ 8.0 & .815 $\pm$ .041 \\
        \tt{Test\_07} & 80 & 59.1 $\pm$ 7.9 & .805 $\pm$ .041 \\
        \tt{Test\_08} & 80 & 60.4 $\pm$ 6.0 & .812 $\pm$ .021 \\
        \tt{Test\_09} & 100 & 56.7 $\pm$ 10.3 & .802 $\pm$ .035 \\
        \tt{Test\_10} & 100 & 54.7 $\pm$ 3.9 & .801 $\pm$ .031 \\
        \bottomrule
    \end{tabular}
    \caption{Performance of model Phase-III-200 in small environments. Generally, it performs worse than the scale-specific models in \cref{tab:main-result}.}
    \label{tab:generalization_down}
\end{table}

%% file: src/sec5-conclusion.tex
\section{Conclusion}\label{sec:conclusion}
We provided a new RL solution to the Flatland3 challenge and achieved a score 4x better than the best RL method before. The key reasons behind the improvement are 1) the tree features and TreeLSTM we adopt and 2) the 20x faster feature parser, which enables us to train our model with far more data than the RL methods before. However, there is still a gap between our method and state-of-the-art OR methods \citep{LiICAPS21}. Our method also takes longer time than OR methods. Another drawback is that there lacks a single model that is able to handle environments of any scale. To achieve optimal performance, we have to train multiple scale-specific models.

%% file: src/appendix.tex
\input{image/cases/cases}

%% file: image/cases/cases.tex
\begin{figure*}[!htb]
    \centering
    \includegraphics[width=\textwidth]{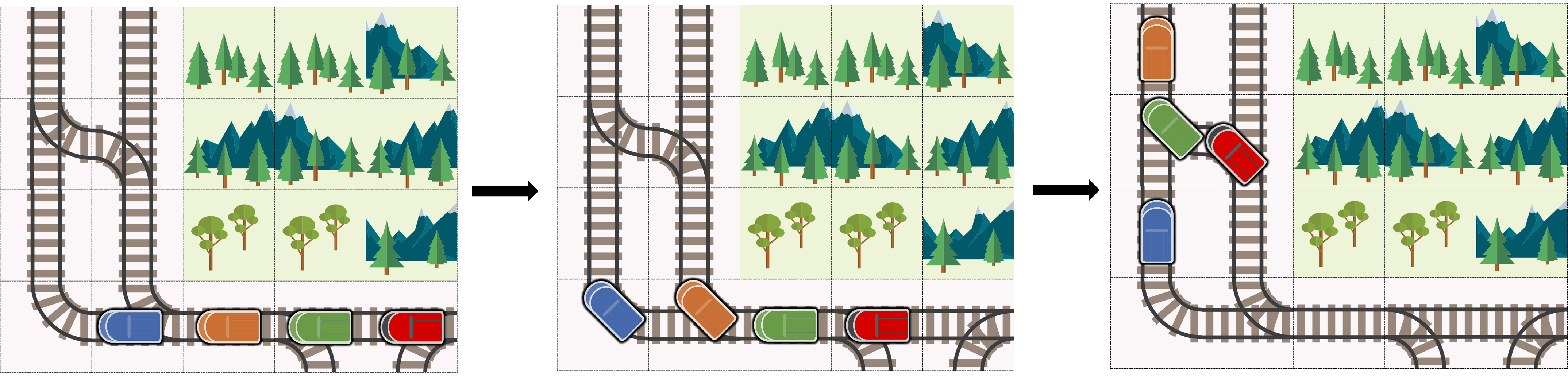}
    \Large
    $T$ \hspace{0.33\textwidth} $T+4$ \hspace{0.27\textwidth} $T+8$
    \caption{Fast trains overtake the slow ones. The speed of the blue train is 0.25 while the other three are 1.0. Fast ones overtake the slow ones to reach targets earlier.}
    \label{fig:overtake}
\end{figure*}

% \begin{figure}[!htb]
%     \centering
%     \includegraphics[width=0.4\textwidth]{image/cases/lineup.png}
%     \caption{Trains line up in a row.}
%     \label{fig:lineup}
% \end{figure}

% \begin{figure}[!htb]
%     \centering
%     \includegraphics[width=0.4\textwidth]{image/cases/parallel.png}
%     \caption{When there are two parallel rail lanes, trains spontaneously line up as if they are in a two-way street.}
%     \label{fig:parallel}
% \end{figure}

\begin{figure*}[!htb]
    \centering
    % \begin{minipage}[t]{0.3\textwidth}
    %     \includegraphics[width=\textwidth]{image/cases/makeway.png}
    %     \caption{Slow train makes way for fast trains. The orange one has a speed of 1.0, and the blue one has a speed of 0.5. The blue one waits until the orange one passes.}
    %     \label{fig:makeway}
    % \end{minipage}
    \begin{minipage}[t]{0.4\textwidth}
        \includegraphics[width=\textwidth]{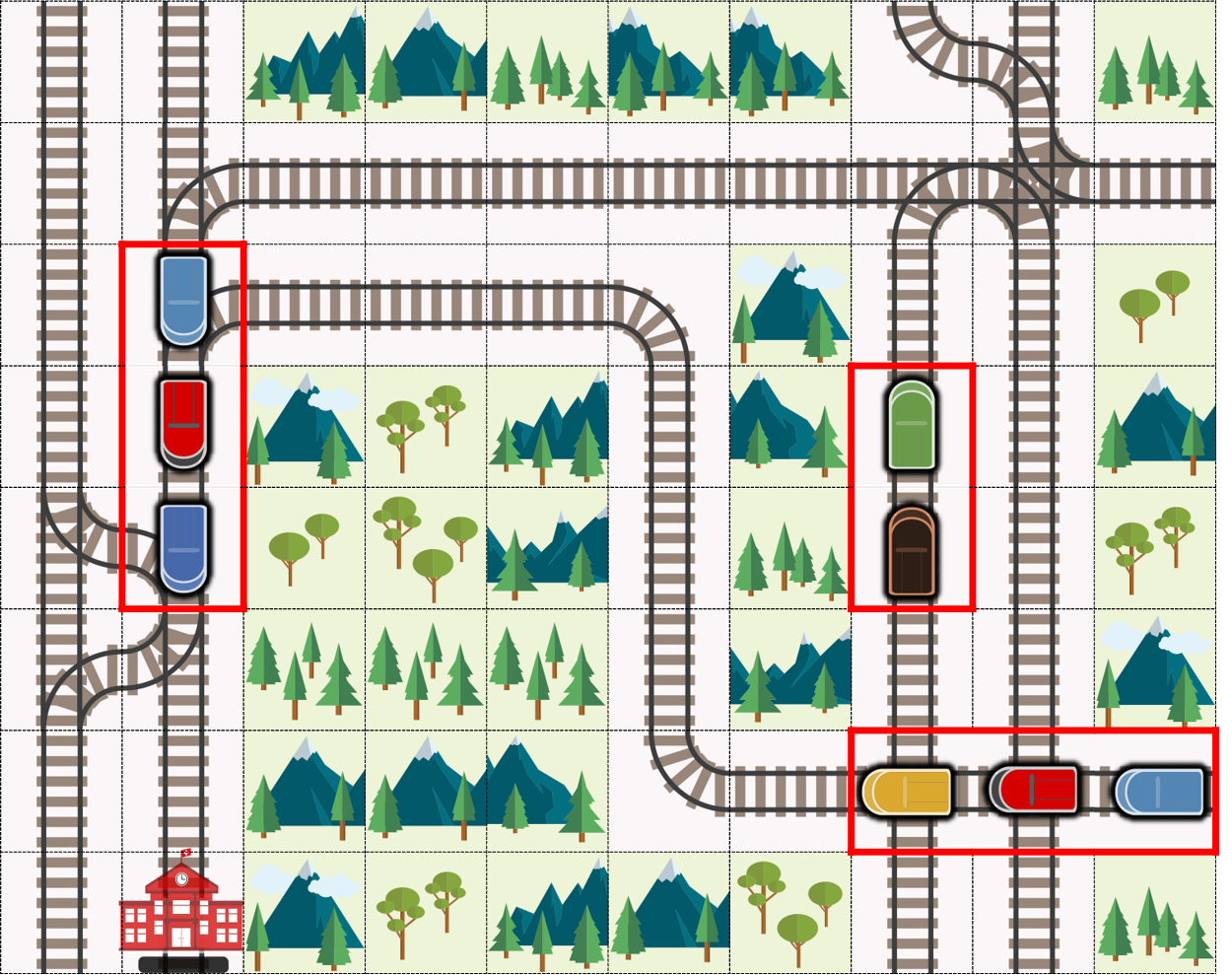}
        \caption{Trains line up in a row.}
        \label{fig:lineup}
    \end{minipage}
    \qquad\qquad
    \begin{minipage}[t]{0.4\textwidth}
        \includegraphics[width=\textwidth]{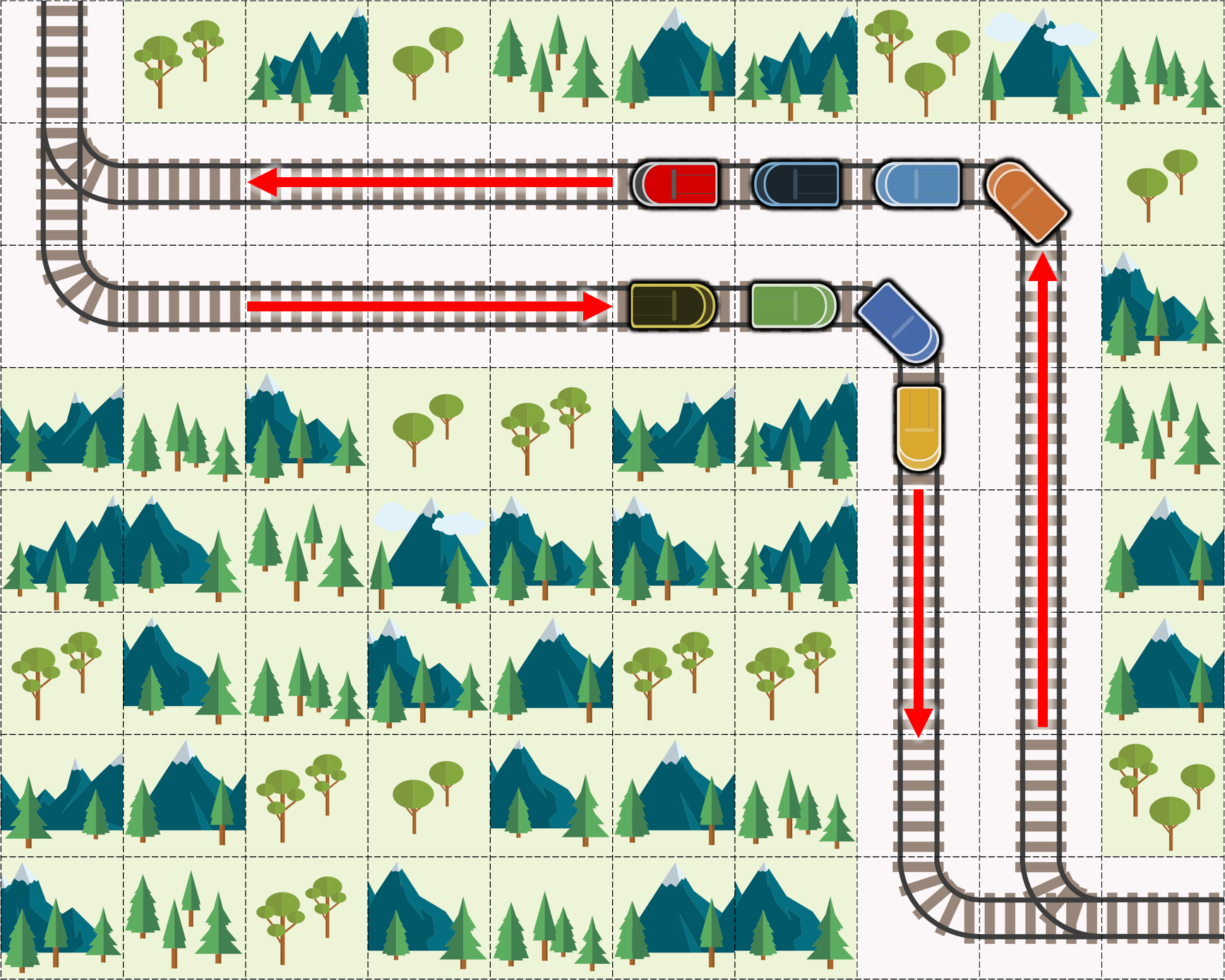}
        \caption{When there are two parallel rail lanes, trains spontaneously line up as if they are in a two-way street.}
        \label{fig:parallel}
    \end{minipage}
\end{figure*}

\begin{figure*}[!htb]
    \centering
    \begin{minipage}[t]{0.5\textwidth}
        \centering
        \includegraphics[width=0.7\textwidth]{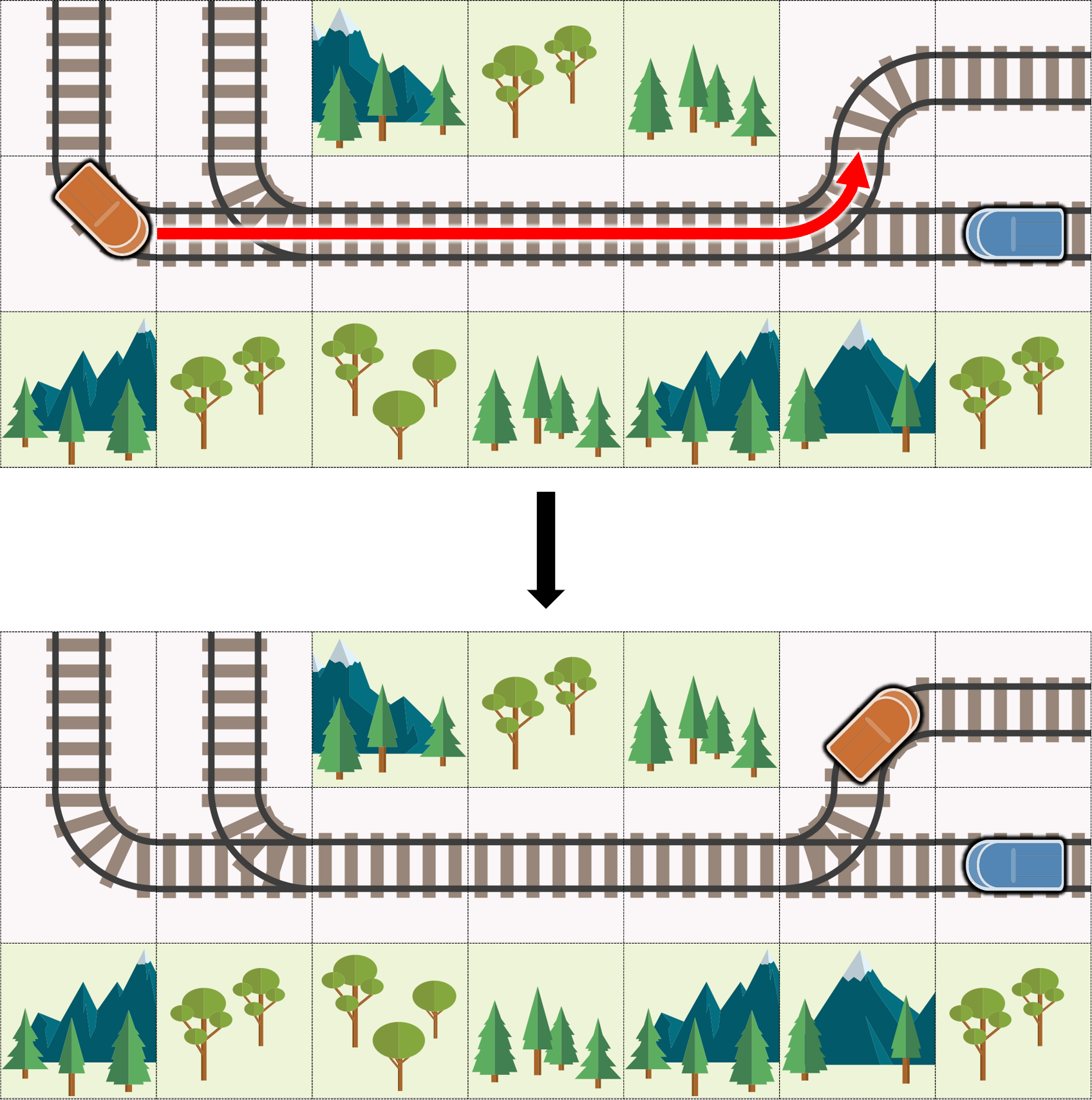}
        \caption{A slow train makes way for a fast one. The orange one has a speed of 1.0, and the blue one has a speed of 0.5. The blue one waits until the orange one passes.}
        \label{fig:makeway}
    \end{minipage}
\end{figure*}